# Variational Temporal Deconfounder for Individualized Treatment Effect Estimation from Longitudinal Observational Data

Zheng Feng[†]
Department of Health Outcomes
and Biomedical Informatics
University of Florida
Gainesville, Florida, USA
fengzheng@ufl.edu

Mattia Prosperi
Department of Epidemiology
University of Florida
Gainesville, Florida, USA
m.prosperi@ufl.edu

Jiang Bian
Department of Health Outcomes
and Biomedical Informatics
University of Florida
Gainesville, Florida, USA
bianjiang@ufl.edu

## ABSTRACT

Estimating treatment effects, especially individualized treatment effects (ITE), using observational data is challenging due to the complex situations of confounding bias. Existing approaches for estimating treatment effects from longitudinal observational data are usually built upon a strong assumption of "*unconfoundedness*", which is hard to fulfill in real-world practice. In this paper, we propose the Variational Temporal Deconfounder (VTD), an approach that leverages deep variational embeddings in the longitudinal setting using proxies (i.e., surrogate variables that serve for unobservable variables). Specifically, VTD leverages observed proxies to learn a hidden embedding that reflects the true hidden confounders in the observational data. As such, our VTD method does not rely on the "unconfoundedness" assumption. We test our VTD method on both synthetic and real-world clinical data, and the results show that our approach is effective when hidden confounding is the leading bias compared to other existing models.

## 1 Introduction

Treatment effect, the causal effect of a given treatment or intervention on an outcome, estimation plays an important role in evidence-based clinical medicine, providing quantified measurements of benefit or harm for the treatment of interest, which help regulators to make regulatory decisions, health care community to develop guidelines and decision support tools, and clinical professionals to decide the treatment choices in their clinical practice. Randomized controlled trials (RCTs) have been widely used to estimate the average treatment effects (ATE), measuring the difference in average outcomes between individuals in the treatment group and those in the control group. In a well-designed RCT, patients are randomly assigned to the control and treatment groups, such that the units in the treatment vs. control groups are identical across all known and unknown factors to reduce the potential bias [11]. However, the inherent limitations of RCTs, i.e., resource-intensive and narrow inclusion criteria, make it not only time-consuming and logistically complex to conduct, but also the study results may not generalize beyond the study population [21]. In recent years, the rapid growth of electronic health record (EHR) systems has made large collections of longitudinal observational real-world data (RWD) available for research to generate real-world evidence (RWE) [23]. Further, there is a strong desire to obtain an individualized treatment effect (ITE), considering the heterogeneity of the target patient population and their differential responses to the same treatment. In recent years, ITE estimation based on more accessible observational data like EHRs has been a thriving research area to fill the gap [2].

One of the most critical issues of estimating ITEs from observational data is confounding - when variables can affect both the outcomes and treatment interventions [12]. These variables are thus called confounders. For example, socioeconomic status can affect both the medication a patient has access to and the patient's general health. Therefore, socioeconomic status acts as a confounder between medication and health outcomes. If confounders can be measured, the most common way to counter their effect is by "*controlling*" them [20]. Many approaches for estimating ITE from observational data have been proposed according to this solution, which can be categorized into 2 groups: (1) covariate adjustment [5, 26–28], and (2) propensity score re-weighting [6, 18, 29]. Most of these approaches are built on the commonly used assumption of "*unconfoundedness,*" where all variables that affect both the interventions and outcomes are observed and measured. However, the *unconfoundedness*-based models will lead to biased ITE estimation when certain confounders are hidden or unmeasured [10]. In reality, it is unlikely that we can observe and directly measure all confounders in real-world observational studies. For example, RWD like EHRs often do not have variables such as environmental factors or personal preferences, which are potential confounders. A possible way of modeling hidden confounders is through modeling their proxies (surrogate measures). For example, we may not have a variable in EHRs to directly measure patients' socioeconomic status. Still, we can use variables such as zip code and employment status in EHRs as the proxies to infer patient socioeconomic status.

Several approaches have built deconfounding ITE estimation models by using proxy variables. The multiplicity of causes [30] and matrix factorization [13] are used to infer the confounders from missing or proxy variables. More recently, the Variational autoencoder (VAE) [17]—a deep generative model with powerful hidden representation learning ability, has been applied to model hidden confounders [15, 20, 31] and achieved superior performance. However, these variational generative model-based approaches are designed for cross-sectional settings and cannot be adopted for a longitudinal setting. In real-world clinical practice, EHRs contain rich time-dependent patient information such as lab results, vital signs, and medications across their encounters with the health system. With these data, we can answer some essential questions, what is the optimal time to administer a treatment, when the treatment regime needs to be stopped, or in which order treatments should be given to obtain the best patient response [2]. Only a few attempts have built longitudinal ITE models [3, 19], and none of them have tried to use the variational generative approach to model the hidden confounders over time.

In this paper, we propose the Variational Temporal Deconfounder (VTD), a novel method for ITE estimation that leverages the autoencoded variational inference to address hidden confounding in a longitudinal setting. Instead of assuming no unobserved hidden confounders, we create embeddings of latent variables to recover the distributions of hidden confounders from the proxies over the observational data space. Our approach is two-fold: (1) a Recurrent Neural Network (RNN)-based factor model that can infer the latent random variables with a variational autoencoder to learn the hidden confounders from variations of observed proxies while capturing the dependencies among the hidden confounders at neighboring timesteps; and (2) a timestep-wise variational lower bound together with the prediction loss to integrate joint training of the latent factor model with the ITE estimation task. We highlight our VTD, same as the time series deconfounder [3], works as an unbiased ITE estimation approach requiring weaker assumptions than standard methods over observational data. To show the effectiveness of VTD, we first conducted a simulation study to investigate VTD's capability to infer latent variables where we explicitly created hidden confounding. Then we evaluate VTD on a real-world dataset (i.e., MIMIC-III) with patients admitted into Intensive Care Units (ICU).

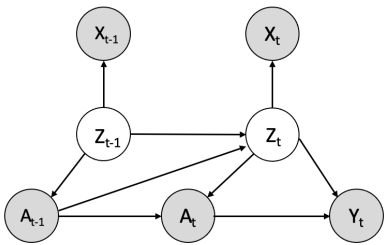

**Figure 1. Graphical factor model of proposed VTD.**

## 2 Methods

### 2.1 Problem Formulation

Let random variables $\mathbf{X}_t^{(i)} = \left[x_{t,1}^{(i)}, x_{t,2}^{(i)}, ..., x_{t,p}^{(i)}\right] \in \mathcal{X}_t$ denote observed p dimensional time-dependent covariates of a patient (i) with time-stamp $t = 1, ..., T$ and $i = 1, ..., N$. Let random variables $\mathbf{A}^{(i)} = \left[a_1^{(i)}, a_2^{(i)}, ..., a_t^{(i)}\right] \in \mathbf{a}_t$ denote a binary treatment assignment, and $\mathbf{Y}^{(i)} = \left[y_1^{(i)}, y_2^{(i)}, ..., y_t^{(i)}\right] \in \mathcal{Y}_t$ be observed outcomes over time-stamp t. For simplicity, we include static features as part of observed covariates $\mathbf{X}$ since it does not change our assumptions. For patient $i = 1, ..., N$, and across time stamp $t = 1, ..., T$, we denote an observed dataset as $\mathcal{D} = \left(\left\{\mathbf{x}_t^{(i)}, \mathbf{a}_t^{(i)}, \mathbf{y}_t^{(i)}\right\}_{t=1}^T\right)_{i=1}^N$. We emphasize that observed covariates $\mathbf{x}_t^{(i)}$ in $\mathcal{D}$ are proxies of true confounders. We denote the true unobserved confounders for proxies $\mathbf{x}_t^{(i)}$ as r-dimensional random variables $\mathbf{Z}$ where $\mathbf{Z}_t^{(i)} = \left[Z_{t,1}^{(i)}, Z_{t,2}^{(i)}, ..., Z_{t,r}^{(i)}\right] \in \mathcal{Z}_t$. Figure 1 shows the causal structure between hidden confounders $\mathbf{Z}$ and other variables.

We adopt the potential outcome framework under the longitudinal setting proposed by Robins and Hernán [24], who extended it from the static setting of Neyman [22] and Rubin [25]. Let $\overline{(\cdot)}_t$ denote the historical covariates collected before time t. For each patient, given observed covariates $\overline{\mathbf{X}} = [x_1, x_2, ..., x_t] \in \bar{x}_t$ and treatment of $\overline{\mathbf{A}} = [a_1, a_2, ..., a_t] \in \bar{a}_t$, we want to estimate individualized treatment effects (ITE), i.e., potential outcomes $\mathbf{Y}_{(\bar{a})}$ conditioned on $\overline{\mathbf{A}}_{t-1}, \overline{\mathbf{X}}_t$ as

$$\mathbb{E}\left[\mathbf{Y}_{(\bar{a}\geq t)} \mid \overline{\mathbf{A}}_{t-1}, \overline{\mathbf{X}}_t\right] \quad (1)$$

We adopt two standard assumptions [9] for ITE estimation:

**Assumption 1. Consistency.** *If $A_t = a_t$, then the potential outcome for treatment assignment $a_t$ is the same as the observed outcome, i.e., $Y_{t+1}[a_t] = Y_{t+1}$.*

**Assumption 2. Positivity (Overlap).** *If $\mathbb{P}(\overline{A}_{t-1} = \bar{a}_{t-1}, \overline{X}_t = \bar{x}_t) \neq 0$ then $\mathbb{P}(A_t = a_t \mid \overline{A}_{t-1} = \bar{a}_{t-1}, \overline{X}_t = \bar{x}_t) > 0$ for all $a_t$*

Other than these two assumptions, the majority of other methods also assume unconfoundedness or sequential ignorability, i. e.,

$$\mathbf{Y}_{t+1}[\mathbf{a}_t] \perp\!\!\!\perp \mathbf{A}_t \mid \overline{\mathbf{A}}_{t-1}, \overline{\mathbf{X}}_t \quad (2)$$

for all $\mathbf{a}_t \in \mathcal{A}_t$ and $t \in \{0, ..., T\}$, which holds only if there are no hidden confounders. In our setting, we observe proxies $\overline{\mathbf{X}}_t$ instead of true confounders $\overline{\mathbf{Z}}_t$, where unconfoundedness is violated, and using standard methods will result in biased ITE estimation.

We address this by using the VTD, which learns a hidden embedding that reflects the true hidden confounders $\mathbf{Z}_t$ from variations of observed proxies $\mathbf{X}_t$ and also captures the dependencies among $\mathbf{Z}_t$ at neighboring timesteps.

### 2.2 Variational Temporal Deconfounder (VTD)

We introduce our VTD as follows: (1) the architecture of the VTD that consists of a variational RNN autoencoder and an ITE block to produce the hidden embedding and ITE estimation, respectively; and (2) the variational bound of VTD, which ensures the embedding of the hidden confounders can be learned by standard gradient-based optimization.

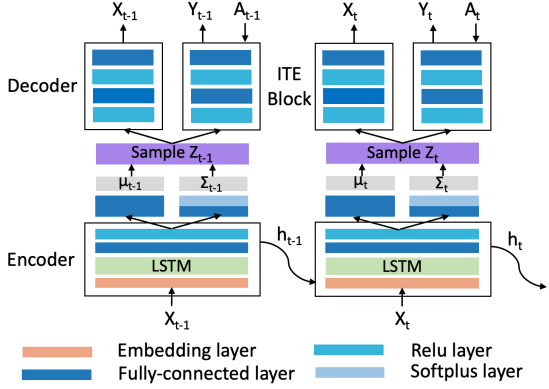

Figure 2: Proposed VTD architecture. $Z_t$ is generated by $\varphi_{enc}$ which takes the hidden states of RNN $h_{t-1}$ and $h_t$ as the input. $(\mu_t, \Sigma_t)$ is generated from function $g$ which takes $z_t$ as the input.

*2.2.1 Architecture of the VTD.* The VTD consists of two main components: (1) a variational RNN autoencoder, which learns the latent variables of hidden confounders $z_t$ the observed proxies $x_t$; and (2) an ITE estimation block, which takes learned hidden embedding $\hat{z}_t$ to predict the probability of receiving treatment $\hat{a}_t$ and potential outcome $\hat{y}_t$. We illustrate the architecture of VTD in Figure 2.

(1) The variational RNN autoencoder. The variational RNN autoencoder uses an RNN Encoder-Decoder framework where the RNN is introduced to adjust $z_t$ for the time-varying structure of a longitudinal setting shown in Figure 1. The encoder maps the input proxies $x_t$ from the observed space to the latent space of hidden confounders $z_t$. In the encoder, an RNN layer takes a sequence of observed proxies x and outputs the hidden states h accordingly. A followed fully-connected layer $\Phi_{enc}$ takes outputs of the RNN layer to map the hidden states $h_t$ and $h_{t-1}$ of each time step $t$ onto the latent embedding $\hat{z}_t$, i.e.,

$$\hat{z}_t = \Phi_{enc}(h_{t-1}, h_t) \quad (3)$$

Then the decoder takes the embedding $\hat{z}_t$ as input and generates the proxies $x_t$ with a variational mapping function $g$, which takes mapping from $\hat{z}_t$ and $h_{t-1}$ to parameters vectors $\mu$ and $\Sigma$, and then generate samples as the input of the decoder network $\Phi_{dec}$

$$\hat{x}_t = \Phi_{dec}(g(\hat{z}_t)) \quad (4)$$

(2) The ITE estimation block. Leveraging the learned hidden embedding $\hat{z}_t$ as the confounding representation, we estimate the ITE by incorporating two tasks of predicting the probability of receiving treatment $\hat{w}_t$ and the outcome $\hat{y}_t$.

We use a fully-connected layer $f_a$ that takes the embedding $\hat{z}_t$ to predict the predicted probability of receiving treatment as $\hat{a}_t$, i.e.,

$$\hat{a}_{t,} = f_a(\hat{z}_t) \quad (5)$$

We also use a fully-connected layer $f_y$ to predict the outcome $\hat{y}_t$, which takes the hidden embedding $\hat{z}_t$ together with the assigned treatments $a_t$, i. e.,

$$\hat{y}_{t,a_t} = f_y(\hat{z}_t, a_t) \quad (6)$$

Then we compute the weights using the inverse probability of treatment weighting (IPTW) and extend them to a dynamic setting as follows,

$$\hat{w}_t = \frac{\Pr(A)}{\hat{a}_t} + \frac{(1 - \Pr(A))}{(1 - \hat{a}_t)} \quad (7)$$

where $\Pr(A)$ denotes the probability of being in the treated group. By incorporating (7) with our outcome prediction, we define the supervised loss $L_s$ as

$$L_s = E[\hat{w}_t(\hat{y}_t - y_t)] \quad (8)$$

*2.2.2 The variational bound of VTD.* The VAE was proposed to model complex multimodal distributions of hidden factors over the space of the observed dataset. We define the joint distribution of observed proxies x and latent confounders z over $T$ time steps as follows,

$$p(x_{\leq T}, z_{\leq T}) = \prod_{t=1}^{T} p(x_t \mid z_{\leq t}, x_{<t}) p(z_t) \quad (9)$$

In the standard VAE, the latent random variable $z$ follows a standard Gaussian distribution. To reflect the causal structure in Figure 1, we assume $z_t$ follows a prior Gaussian distribution as

$$z_t \sim \mathcal{N}(\mu_t, \Sigma_t), \text{where } [\mu_t, \Sigma_t] = f(h_t, h_{t-1}) \quad (10)$$

where $f$ is a function that maps the hidden states of RNN $h_{t-1}$ and $h_t$ to the parameter space of $\mu_t$ and $\Sigma_t$. And we also assume $x_t \mid z_t$ follows a Gaussian distribution

$$x_t \mid z_t \sim \mathcal{N}(\mu_t, \Sigma_t), \text{where } [\mu_t, \Sigma_t] = g(z_t) \quad (11)$$

Now our goal is to infer the parameter of the posterior $p(z_{\leq T} \mid x_{\leq T})$. By following the paradigm in [4, 16] we introduce the variational distribution $q(z_t \mid x_{\leq t}, z_{<t})$ and transfer the problem of inferencing $p(z_{\leq T} \mid x_{\leq T})$ to maximize

$$L_{ELBO} = \mathbb{E}_{q(z_{\leq t}, x_{<t})} \left[ \sum_{t=1}^{T} (\log p(x_t \mid z_{\leq t}, x_{<t}) - KL(q(z_t \mid x_{\leq t}, z_{<t}) \| p(z_t \mid x_{<t}, z_{<t}))) \right] \quad (12)$$

where $L_{ELBO}$ denotes the marginal likelihood lower bound (ELBO) of the full dataset. We incorporate the supervised loss of ITE estimation $L_s$ and $L_{ELBO}$ to define loss $L$ as

$$L = L_s - \alpha L_{ELBO} \quad (13)$$

where $\alpha$ controls the effects of hidden confounding. We jointly train the variational RNN autoencoder and ITE estimation block by optimizing **L**.

## 3 Experiments

We demonstrate the effectiveness of the VTD in experiments using both 1) synthetic and 2) real-world data. We show that the VTD reduces confounding bias in ITE estimation from the empirical observation from both experiments. We compared VTD with the following causal inference approaches (1) G-formula, a generalized approach to the standard regression model over the longitudinal setting that can be used to adjust for time-varying confounders [14]; (2) Deep Sequential Weighting (DSW), which infers the hidden confounders by incorporating the current treatment assignments and historical information using a deep recurrent weighting neural network [19]; and (3) Time Series Deconfounder (TSD), which leverages the assignment of multiple treatments over time to enable the estimation of treatment effects in the presence of multi-cause hidden confounders [3].

We report the Rooted Mean Square Error (RMSE) between predicted and ground truth outcomes to measure models' performance on conventional prediction tasks. To evaluate ITE estimation, the most common measurement is the Precision in Estimation of Heterogenous Effect (PEHE) [7], defined as the mean squared error between the ground truth and estimated ITE, i.e.,

$$\text{PEHE} = \frac{1}{N}\sum_{i=1}^{N}\left(\left(y_1^{(i)} - y_0^{(i)}\right) - \left(\hat{y}_1^{(i)} - \hat{y}_0^{(i)}\right)\right)^2 \quad (14)$$

However, in real-world, the counterfactual is never observed; thus, we use the influence function - PEHE (IF-PEHE) that approximates the true PEHE by "*derivatives*" of the PEHE function [1].

**Table 1: Performance comparison on synthetic datasets. Here we report RMSE and IF-PEHE.**

| Model | RMSE | IF-PEHE |
|---|---|---|
| G-formula | 5.46 ± 0.11 | 30.42 ± 4.64 |
| DSW | 2.63 ± 0.05 | 10.28 ± 1.06 |
| TSD | 3.06 ± 0.14 | 23.65 ± 2.23 |
| VTD (ours) | **2.46 ± 0.10** | **9.86 ± 1.21** |

### 3.1 Experiments with synthetic data

Following the setting of Bica et al [3], we construct a synthetic dataset with longitudinal variables and hidden confounders, consisting of 4000 samples over 10 time steps. Each individual has 5 hidden confounders and 100 observed covariates. All reported results are averaged on 30 realizations and give the corresponding standard deviations.

As shown in Table 1, our VTD model outperforms all other models on both RMSE and IF-PEHE on the synthetic data. This shows variational embedding of VTD can better capture the information of hidden confounders over temporal structure compared with other approaches thus achieving a better estimation of ITE. And we also observe that deep representation-based models achieve better performance over the baseline G-formula by a large margin, which can be attributed to their ability to deal with complex and high dimensional data by using neural networks as the backbone.

**Table 2: Performance comparison on real-world datasets. Here we report RMSE and IF-PEHE.**

| Model | Vasopressor-Meanbp | | Ventilator-SpO2 | |
|---|---|---|---|---|
| | RMSE | IF-PEHE | RMSE | IF-PEHE |
| G-formula | 12.53 ± 0.27 | 63.35 ± 5.43 | 1.57 ± 0.14 | 53.28 ± 5.21 |
| DSW | **8.55 ± 0.22** | **12.01 ± 2.33** | **1.06 ± 0.11** | **8.68 ± 1.54** |
| TSD | 9.34 ± 0.10 | 57.26 ± 4.71 | 1.23 ± 0.07 | 35.21 ± 4.85 |
| VTD (ours) | 9.13 ± 0.23 | 24.16 ± 3.62 | 1.20 ± 0.11 | 19.27 ± 2.71 |

### 3.2 Experiments with Real-world data: MIMIC-III

To evaluate the performance of VTD in a real-world application, we adopt a setting based on the MIMIC-III dataset [19], where we constructed a dataset with 11,715 ICU patients. We considered 2 different treatment assignments: vasopressors and mechanical ventilator, and evaluate their causal effect on important outcomes, separately; for vasopressors, we use mean blood pressure (Meanbp), while for mechanical ventilator, we adopt oxygen saturation (SpO2) as the outcomes following previous work. We considered 27 time-varying covariates and 12 static demographics. Table 2 shows the performance of 4 models on both Vasopressor-Meanbp and Ventilator-SpO2 tasks.

## 4 Discussion and conclusion

In this paper, we introduced a novel approach, VTD, for estimating ITE in a longitudinal setting by using deep variational generative models to address hidden confounders over time. Our experiments using synthetic data have shown that our approach is more effective when hidden confounding is the leading bias compared to other existing models. In the next step, we will test approaches with a different number of confounders to evaluate their robustness. In the real-world application using MIMIC-III dataset, we can see VTD consistently performs better than the baseline G-formula and the TSD model, however, worse – but competitive – than DSW on both the RMSE and IF-PEHE. However, DSW is a deep learning-based approach built on the assumption of unconfoundedness, which VTD does not assume. This may indicate that in some real-world scenarios, the leading bias of ITE may not come from the hidden confounders. The approaches of controlling bias by directly modelling the hidden confounders may not be the most suitable choice. Nevertheless, further investigations are needed as the unconfoundedness assumption may not hold in certain real-world applications. Identifying the types of real-world applications where unconfoundedness holds or not is thus critical to guide the choice of the modeling approach.

## ACKNOWLEDGMENTS

This work was in part supported by the NIH Awards R01CA2464-18, R01AG076234, R21AG068717, and 3R21ES032762-02S1.